\documentclass[acmtog]{acmart}
\acmSubmissionID{251}

\usepackage{subcaption}
\AtBeginDocument{%
  \providecommand\BibTeX{{%
    \normalfont B\kern-0.5em{\scshape i\kern-0.25em b}\kern-0.8em\TeX}}}
    
\makeatletter
\renewcommand*\env@matrix[1][\arraystretch]{%
  \edef\arraystretch{#1}%
  \hskip -\arraycolsep
  \let\@ifnextchar\new@ifnextchar
  \array{*\c@MaxMatrixCols c}}
\makeatother

\setcopyright{acmcopyright}
\copyrightyear{2022}
\acmYear{2022}
\acmDOI{XXXXXXX.XXXXXXX}

\begin{document}

%%
%% The "title" command has an optional parameter,
%% allowing the author to define a "short title" to be used in page headers.
\title{Analytically Integratable Zero-restlength Springs for Capturing Dynamic Modes unrepresented by Quasistatic Neural Networks}

%%
%% The "author" command and its associated commands are used to define
%% the authors and their affiliations.
%% Of note is the shared affiliation of the first two authors, and the
%% "authornote" and "authornotemark" commands
%% used to denote shared contribution to the research.
\author{Yongxu Jin}
\email{yxjin@stanford.edu}
\affiliation{%
  \institution{Stanford University}
  \city{Stanford}
  \state{California}
  \country{USA}
}
\affiliation{%
  \institution{Epic Games}
  \city{Cary}
  \state{North Carolina}
  \country{USA}
}

\author{Yushan Han}
\email{yushanh1@math.ucla.edu}
\affiliation{%
  \institution{University of California, Los Angeles}
  \city{Los Angeles}
  \state{California}
  \country{USA}}
\affiliation{%
  \institution{Epic Games}
  \city{Cary}
  \state{North Carolina}
  \country{USA}
}

\author{Zhenglin Geng}
\email{zhenglin.geng@epicgames.com}
\affiliation{%
  \institution{Epic Games}
  \city{Cary}
  \state{North Carolina}
  \country{USA}
}

\author{Joseph Teran}
\email{jteran@math.ucdavis.edu}
\affiliation{%
  \institution{University of California, Davis}
  \city{Davis}
  \state{California}
  \country{USA}}
\affiliation{%
  \institution{University of California, Los Angeles}
  \city{Los Angeles}
  \state{California}
  \country{USA}}
\affiliation{%
  \institution{Epic Games}
  \city{Cary}
  \state{North Carolina}
  \country{USA}
}

\author{Ronald Fedkiw}
\email{rfedkiw@stanford.edu}
\affiliation{%
  \institution{Stanford University}
  \city{Stanford}
  \state{California}
  \country{USA}
}
\affiliation{%
  \institution{Epic Games}
  \city{Cary}
  \state{North Carolina}
  \country{USA}
}

%%
%% By default, the full list of authors will be used in the page
%% headers. Often, this list is too long, and will overlap
%% other information printed in the page headers. This command allows
%% the author to define a more concise list
%% of authors' names for this purpose.
% \renewcommand{\shortauthors}{Trovato and Tobin, et al.}

%%
%% The abstract is a short summary of the work to be presented in the
%% article.
\begin{abstract}
We present a novel paradigm for modeling certain types of dynamic simulation in real-time with the aid of neural networks. 
In order to significantly reduce the requirements on data (especially time-dependent data), as well as decrease generalization error, our approach utilizes a data-driven neural network \textit{only} to capture quasistatic information (instead of dynamic or time-dependent information). 
Subsequently, we augment our quasistatic neural network (QNN) inference with a (real-time) dynamic simulation layer.
Our key insight is that the dynamic modes lost when using a QNN approximation can be captured with a quite simple (and decoupled) zero-restlength spring model, which can be integrated analytically (as opposed to numerically) and thus has no time-step stability restrictions.
Additionally, we demonstrate that the spring constitutive parameters can be robustly learned from a surprisingly small amount of dynamic simulation data.
Although we illustrate the efficacy of our approach by considering soft-tissue dynamics on animated human bodies, the paradigm is extensible to many different simulation frameworks. 
\end{abstract}

%%
%% The code below is generated by the tool at http://dl.acm.org/ccs.cfm.
%% Please copy and paste the code instead of the example below.
%%


%%
%% Keywords. The author(s) should pick words that accurately describe
%% the work being presented. Separate the keywords with commas.

\begin{teaserfigure}
    \includegraphics[width=\textwidth]{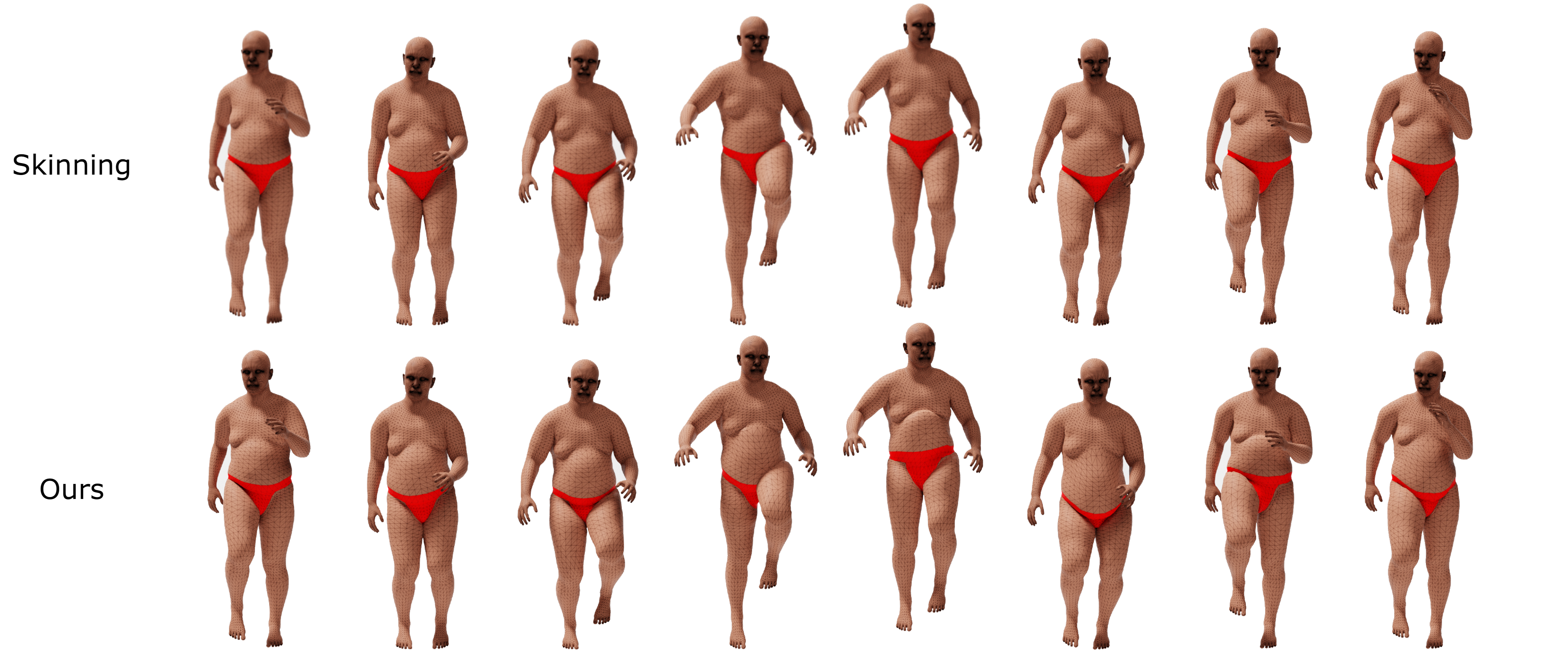}
    \caption{Our method enhances standard skinning with a configuration-only quasistatic neural network (QNN) that approximates quasistatic hyperelasticity as well as analytically integratable zero-restlength springs trained that approximates inertial effects. The QNN fixes well-known skinning artifacts (e.g. in the shoulder regions) and the zero-restlength springs add ballistic motion (e.g. in the belly region). We refer readers to our supplementary video which is far more compelling than still images.}
\end{teaserfigure}

%%
%% This command processes the author and affiliation and title
%% information and builds the first part of the formatted document.
\maketitle

\section{Introduction}
\label{sec:intro}

Recently, there has been a lot of interest in using neural networks to approximate dynamic simulation (see e.g. \cite{holden2019subspace, pfaff2020learning, sanchez2020learning, santesteban2020softsmpl}), especially because neural network inference has the potential to run in real-time (on high-end GPUs). Unfortunately, one requires an exorbitant amount of training data in order to represent all the possible temporal transitions between states that these networks aim to model. These networks do not typically generalize well when not enough training data is used. Even if one had access to the exorbitant amount of training data required, an unwieldy amount of network parameters would be required to prevent underfitting.

Some aspects of a dynamic simulation depend mostly on the configuration, whereas others more strongly depend on the time history of configuration to configuration transitions; thus, we propose the following paradigm. 
Firstly, we construct a neural network that depends only on the configurations (and as such cannot capture dynamic modes). Secondly, we subtract this configuration-only model from the full dynamics in order to obtain a dynamics layer. Thirdly, we propose a dynamic simulation model that can approximate the dynamics layer. Theoretically, a well-approximated dynamics layer has the potential to augment the configuration-only neural network in a way that exactly matches the original simulations. Moreover, if the configuration-only neural network can capture enough of the non-linearities, then the dynamics layer has the potential to be quite simple (and thus real-time). 

In this paper, we propose using a quasistatic physics simulation neural network (see e.g. \cite{jin2020pixel, luo2018nnwarp, geng2020coercing, bertiche2020pbns}) as the configuration-only neural network. Since quasistatic neural networks (QNNs) do not have time dependency, they require far less training data and as such can use a much simpler network structure with far fewer parameters than a network that attempts to model temporal transitions.
Using less training data on a network designed to capture temporal transitions leads to overfitting and poor generalization to unseen data. Using a simpler network structure with less parameters on a network designed to capture transitions leads to underfitting of the training data (and poor generalization).

Although we expect that an entire cottage industry could be developed around the modelling and real-time simulation of dynamics layers, we propose only a very simple demonstrational model here (but note that it works surprisingly well). Importantly, the zero-restlength spring approximation to the dynamics layer can be integrated analytically and thus has zero truncation error and no time step stability restrictions, making it quite fast and accurate. Furthermore (as shown in Section \ref{sec:result}), one can (automatically) robustly learn spring constitutive parameters from a very small amount of dynamic simulation data.

\section{Related Work}

\paragraph{Stated-based methods} We first discuss prior works that generate elastic deformation directly from spatial state without considering temporal or configurational history. Many works aim to upsample a low-resolution simulation to higher resolution: \cite{feng2010deformation} trains a regressor to upsample,  \cite{kavan2011physics} learns an upsampling operator, and \cite{chen2018synthesizing} rasterizes the vertex positions into an image before upsampling it and interpolating new vertex positions. \cite{wang2010example, zurdo2012animating, xu2014sensitivity} use example-based methods to synthesize fine-scale wrinkles from a database. \cite{patel2020tailornet} predicts a low-frequency mesh with a fully connected network and uses a mixture model to add wrinkles.  \cite{chentanez2020cloth} upsamples with graph convolutional neural networks. \cite{wu2021recovering} recovers high-frequency geometric details with perturbations of texture.
\cite{lahner2018deepwrinkles} uses a generative adversarial network (GAN) to upsample a cloth normal map for improved rendering.
\cite{bailey2018fast, bailey2020fast} use neural networks to drive fine scale details from a coarse character rig.
Many works aim to learn equilibrium configurations from boundary conditions: 
\cite{luo2018nnwarp} uses a neural network to add non-linearity to a linear elasticity model.
\cite{mendizabal2020simulation} learns the non-linear mapping from contact forces to displacements.
Such approaches are particularly common in virtual surgery applications, e.g. \cite{liu2020real,de2011physics,salehi2021physgnn,roewer2018towards, pfeiffer2019learning}.
\cite{jin2020pixel} trains a CNN to infer a displacement map which adds wrinkles to skinned cloth, and \cite{wu2020skinning} improves the accuracy of this approach by embedding the cloth into a volumetric tetrahedral mesh.
\cite{bertiche2020pbns} adds physics to the loss function, a common approach in physics-inspired neural networks (PINNs), see e.g. \cite{raissi2019physics}.
To avoid the soft constraints of PINNs that only coerce physically-inspired behaviour, \cite{geng2020coercing, srinivasan2021learning} add quasistatic simulation as the final layer of a neural network in order to constrain output to physically attainable manifolds.

\paragraph{Transition-based methods} Here we discuss prior works that use a temporal history of states, typically for resolving dynamic/inertia related behaviors. In one of the earliest works (before the deep learning era) \cite{grzeszczuk1998neuroanimator} uses a neural network to learn temporal transitions and leverage back propagation to optimize control parameters. 
\cite{de2010stable} incorporates an approximation to the quasistatic equilibrium that serves as a control for a dynamics layer.
\cite{guan2012drape} predicts a cloth mesh from body poses and previous frames, solving a linear system to fix penetrations.
\cite{hahn2014subspace} uses dynamic subspace simulation on an adaptive selected basis generated from the current body pose.
\cite{holden2019subspace} computes a linear subspace of configurations with principal component analysis (PCA) and learns subspace simulations from previous frames with a fully connected network. \cite{fulton2019latent, tan2020realtime, tan2018mesh} obtain nonlinear subspaces with autoencoder networks. Similar methods are commonly used to animate fluids using regression forests \cite{ladicky2015data} or recurrent neural networks (RNNs) \cite{wiewel2019latent}.  \cite{pfaff2020learning} and \cite{sanchez2020learning} use graph networks to learn simulations with both fixed and changing topology. \cite{chentanez2020cloth} proposes a transition-based model with position and linear/angular velocity of the body as network input (in addition to a state-based model).
\cite{meister2020deep} uses a fully connected network to predict node-wise acceleration for total Lagrangian explicit dynamics. \cite{deng2020alternating} proposes a convolutional long short-term memory (LSTM) layer to capture elastic force propagation.
\cite{zhang2021dynamic} uses an image based approach to enhance detail in low resolution dynamic simulations.

\paragraph{Secondary dynamics for characters} Numerical methods that resolve the dynamic effects of inertia-driven deformation have a long history in computer graphics skin and flesh animation. We refer interested readers to only a few recent papers and a plethora of references therein (e.g. \cite{wang2020adjustable,Zhang:CompDynamics:2020,sheen2021volume}). We note that any of these techniques could be used to generate training data for learning-based methods. Secondary dynamics for characters have also been added using data-driven methods: \cite{pons2015dyna} provides a motion capture dataset with dynamic surface meshes, and proposes a linear auto-regressive model to capture dynamic displacements compressed by PCA. \cite{loper2015smpl} extends this method to the SMPL human model. 
See also \cite{casas2018learning, santesteban2020softsmpl, seo2021dsnet}. \cite{kim2017data} proposes a two layer approach which skins a volumetric body model as an inner layer and simulates a tetrahedral mesh as an outer layer. The constitutive parameters of the outer layer are learned from 4D scan data. \cite{zheng2021deep} trains a network to approximate per-vertex displacements from temporal one-ring state using backward Euler simulation data of primitive shapes. \cite{deng2020alternating} also uses a one-ring based approach and trains with forward Euler.

\paragraph{Proportional-derivative control} Our analytic zero-restlength spring targeting method resembles proportional-derivative (PD) control algorithms used in both computer graphics and robotics. We refer interested readers to several papers leveraging PD control and control parameter optimization for various usages \cite{allen2011analytic, weinstein2007impulse, wang2012optimizing, hodgins1995animating, de2005pd}.

\section{Quasistatic Neural Network}
\label{sec:qnn}

We use the (freely available) MetaHuman \cite{metahuman} which has 122 joints and 13575 vertices as our human model. Given joint angles $\boldsymbol{\theta}$, we use a skinning function $\mathbf{x}^{skin}(\boldsymbol \theta)$ to get the skinned position for each surface vertex.
Any reasonable skinning approach (e.g. linear blend skinning \cite{magnenat1988joint, lewis2000pose} and dual quaternion skinning \cite{kavan2007skinning}) may be used.

Starting from $\boldsymbol{\theta}$ and $\mathbf{x}^{skin}(\boldsymbol \theta)$, we aim to train a neural network that predicts a more realistic surface mesh $\mathbf{x}^{net}(\boldsymbol \theta)$. Generally speaking, we could add our analytically intergratable zero-restlength springs directly on top of the skinning result (and there are many interesting skinning-related methods being proposed recently, e.g. \cite{wu2021agentdress}), although our proposed dynamics layer (likely) works best when the shape of the surface skin mesh is approximated as accurately as possible. We obtain ground truth for $\mathbf{x}^{net}(\boldsymbol \theta)$ via two different approaches: quasistatic simulation (as discussed in Section 3.1) and 4D scanning (which will be discussed in a future paper). Both approaches worked rather well in our experiences.

\subsection{Quasistatic Simulation}
First, we use Tetgen \cite{si2015tetgen} (alternatively, \cite{hu2018Tetwild},\cite{shewchuk98} could be used) to create a volumetric tetrahedron mesh whose boundary corresponds to the Metahuman surface mesh in a reference A-pose. Next, we interpolate skinning weights from the Metahuman surface vertices to the tetrahedron mesh boundary vertices, and subsequently solve a Poisson equation on the tetrahedron mesh to propagate the skinning weights to interior vertices \cite{cong2015fully}. Then, we use a geometric approximation to a skeleton in order to specify which interior vertices of the tetrahedron mesh should follow their skinned positions with either Dirichlet boundary conditions or zero-restlength spring penalty forces.

Our training dataset includes about 5000 poses genetared randomly, from motion capture data, and manually specified animations. Given any target pose, specified by a set of joint angles $\boldsymbol\theta$,  we solve for the equilibrium configuration of the volumetric tetrahedron mesh using the method from \cite{QS} in order to avoid issues with indefiniteness and the method from \cite{marquez2022} to enforce contact boundary conditions on the surface of the tetrahedron mesh.
Although simulation can be time-consuming, quasistatic simulation is much faster than dynamic simulation. Furthermore, the amount of simulation required is significantly smaller than that which would be needed to obtain similar efficacy for a network aiming to capture temporal information, since such a network would require far more parameters to prevent underfitting.

\subsection{QNN}

\begin{figure}
    \includegraphics[width=0.5\textwidth]{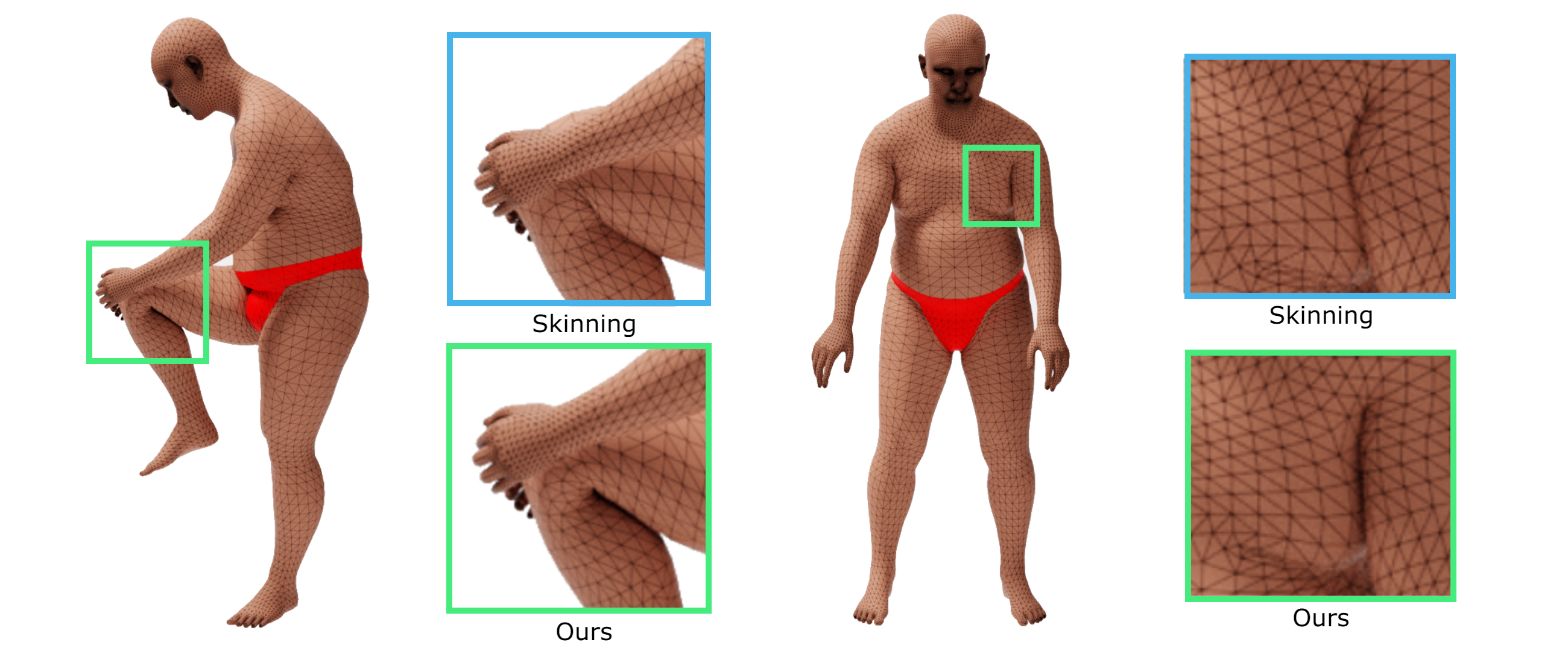}
    \caption{Our QNN resolves well-known skinning collision artifacts. We demonstrate this in extreme poses involving the back of the knee and the armpit.}
    \label{fig:quasi_example}
\end{figure}

Instead of inferring the positions of the surface vertices directly, we augment the skinning result $\mathbf{x}^{skin}(\boldsymbol{\theta})$ with per-vertex displacements $\mathbf{d}(\boldsymbol \theta)$ so that the non-linearities from joint rotations $\boldsymbol \theta$ are mostly captured by the skinning.
This reduces the demands on the neural network allowing for a smaller model and thus requiring less training data.
Given ground truth displacements $\mathbf{d}(\boldsymbol \theta)$, we train our quasistatic neural network (QNN) to minimize the loss between $\mathbf{d}(\boldsymbol \theta)$ and the network inferred result $\mathbf{d}^{net}(\boldsymbol \theta)$. We follow an approach similar to \cite{jin2020pixel} rasterizing the per-vertex displacements into a displacement map image so that a convolutional neural network (CNN) can be used. Of course, one could alternatively use PCA with a fully connected network; however, GPUs are more amenable to the image-based frameworks used by CNNs (see e.g. \cite{wang2021gpu}, which discusses the benefit of using data structures that resemble images on GPUs).
Our QNN can fix skinning artifacts like interpenetration and volume loss (see Figure \ref{fig:quasi_example}), thus providing a simpler dynamics layer for analytic zero-restlength springs to capture (see Section \ref{sec:result} for discussions).

\section{Kinematics}
\label{sec:kinematics}
The skeletal animation will be queried at a user-specified time scale (likely proportional to the frame rate). While these samples are inherently discrete, our approach utilizes the analytic solution of temporal ODEs; therefore, we extend these discrete samples to the continuous time domain. Specifically, given a sequence of skeletal joint angles $\left\{\boldsymbol\theta^1,\boldsymbol\theta^2,\hdots\right\}$, we construct a target function of surface vertex positions $\mathbf{\hat{x}}(t)$ defined for all $t \ge 0$. Options include e.g. Heaviside (discontinous), piecewise linear ($C^0$), or cubic ($C^1$) interpolation. We utilize cubic interpolation given its relative simplicity and favorable continuity. Between sample $n$ at time $t^n$ and sample $n+1$ at time $t^n+\Delta t$, we define
\begin{align}
\mathbf{\hat{x}}(t^n + s \Delta t) &= \hat{\mathbf{q}}^n (s \Delta t)^3 + \hat{\mathbf{a}}^n (s \Delta t)^2 + \hat{\mathbf{b}}^n s \Delta t + \hat{\mathbf{c}}^n \label{eqn:cubic_no_absorb} \\
&= \mathbf{q}^n s^3 + \mathbf{a}^n  s^2 + \mathbf{b}^n  s + \mathbf{c}^n, \label{eqn:cubic_absorb}
\end{align}
where $s \in [0, 1]$ and Equation \ref{eqn:cubic_absorb} absorbs the powers of $\Delta t$ into the non-hatted variables for simplicity of exposition. Enforcing $C^1$ continuity at times $t^n$ and $t^{n+1}$ requires the following position and derivative constraints
\begin{align}
    \begin{bmatrix}
        0 & 0 & 0 & 1 \\
        0 & 0 & 1 & 0 \\
        1 & 1 & 1 & 1 \\
        3 & 2 &1 & 0 
    \end{bmatrix}
    \begin{bmatrix}
        \mathbf{q}^n\\
        \mathbf{a}^n\\
        \mathbf{b}^n \\
        \mathbf{c}^n
    \end{bmatrix}
    &= 
    \begin{bmatrix}
        \mathbf{{x}}^{net}(\boldsymbol{\theta}^n)\\
        \frac{1}{2}(\mathbf{{x}}^{net}(\boldsymbol{\theta}^{n+1}) - \mathbf{{x}}^{net}(\boldsymbol{\theta}^{n-1}))\\
        \mathbf{{x}}^{net}(\boldsymbol{\theta}^{n+1})\\
        \frac{1}{2}(\mathbf{{x}}^{net}(\boldsymbol{\theta}^{n+2}) - \mathbf{{x}}^{net}(\boldsymbol{\theta}^{n}))
    \end{bmatrix}, \label{eqn:cubic_linear_eqn}
\end{align}
which can readily be solved to determine $\mathbf{q}^n, \mathbf{a}^n, \mathbf{b}^n, \mathbf{c}^n$.
Here, $\mathbf{{x}}^{net}(\boldsymbol{\theta}^{n})\\=\mathbf{x}^{skin}(\boldsymbol \theta^n) + \mathbf{d}^{net}(\boldsymbol \theta^n)$ are QNN-inferred surface vertex positions at time $t^n$. Note, in the first interval, $\frac{1}{2}(\mathbf{{x}}^{net}(\boldsymbol{\theta}^{n+1}) - \mathbf{{x}}^{net}(\boldsymbol{\theta}^{n-1}))$ is replaced by the one-sided difference $\mathbf{{x}}^{net}(\boldsymbol{\theta}^{n+1}) - \mathbf{{x}}^{net}(\boldsymbol{\theta}^{n})$.

\section{Dynamics}
\label{sec:dynamics}
We connect a particle (with mass $m$) to each kinematic vertex $\hat{\mathbf{x}}(t^n + s\Delta t)$ using a zero-restlength spring (although other analytically integratable dynamic models could be used).
The position of each simulated particle obeys Hooke's law,
\begin{align}
    \ddot{\mathbf{x}}(t) &= k_s (\hat{\mathbf{x}}(t) - \mathbf{x}(t)) + k_d (\dot{\hat{\mathbf{x}}}(t) - \dot{\mathbf{x}}(t)), \label{eqn:ode}
\end{align}
where $k_s$ and $k_d$ are the spring stiffness and damping (both divided by the mass $m$) respectively.
This equation can be analytically integrated (separately for each particle) to determine a closed form solution, which varies per interval because $\mathbf{q}^n, \mathbf{a}^n, \mathbf{b}^n, \mathbf{c}^n$ vary. Consider one interval $[t^n, t^{n+1}]$ with initial conditions
\begin{align}
    \mathbf{x}^n &= \mathbf{x}(t^{n}) \label{eqn:init_pos}\\
    \dot{\mathbf{x}}^n &= \dot{\mathbf{x}}(t^{n}) \label{eqn:init_vel}
\end{align}
determined from the previous interval; then, the closed form solution in this interval can be written as 
\begin{align}
    \mathbf{x} (t^n + s\Delta t) &= e^{-\frac{k_d}{2} \Delta t s } \mathbf{g} (t^n + s\Delta t) + \mathbf{p}(t^n + s\Delta t) \label{eqn:ode_sol_general}
\end{align}
where $s\in[0,1]$. Here $\mathbf{p} (t^n + s\Delta t)$ is the particular solution associated with the inhomogeneous terms arising from the targets $\hat{\mathbf{x}}(t^n + s\Delta t)$
\begin{align}
    \mathbf{p} (t^n + s\Delta t)  &=  \mathbf{\hat{x}}(t^n + s \Delta t) -   \frac{6\mathbf{q}^n  s 
    + 2 \mathbf{a}^n }{k_s \Delta t^2 } + \frac{6 k_d \mathbf{q}^n }{k_s^2 \Delta t^3} \label{eqn:ode_sol_parti}
\end{align}
The spring is overdamped when $k_d^2 - 4k_s>0$, underdamped when $k_d^2 - 4k_s<0$, and critically damped when $k_d^2 - 4k_s=0$. Defining a (unitless) $\epsilon$ for both the overdamped case, $\epsilon = \frac{\Delta t s}{2}\sqrt{k_d^2 - 4k_s}$, and the underdamped case, $\epsilon = \frac{\Delta t s}{2}\sqrt{4k_s - k_d^2}$, allows us to write
\begin{align}
    \mathbf{g}_{\mathbf{o}}(t^n + s\Delta t) &= \boldsymbol{\gamma}_{\mathbf{1}}^n \frac{e^{\epsilon} + e^{-\epsilon}}{2} +  \boldsymbol{\gamma}_{\mathbf{2}}^n \Delta t s \frac{e^{\epsilon} - e^{-\epsilon}}{2\epsilon} \label{eqn:ode_sol_over}\\
    \mathbf{g}_{\mathbf{u}}(t^n + s\Delta t) &= \boldsymbol{\gamma}_{\mathbf{1}}^n \cos\epsilon + \boldsymbol{\gamma}_{\mathbf{2}}^n \Delta t s \frac{\sin\epsilon}{\epsilon}\label{eqn:ode_sol_under}\\
    \mathbf{g}_{\mathbf{c}}(t^n + s\Delta t) &= \boldsymbol{\gamma}_{\mathbf{1}}^n + \boldsymbol{\gamma}_{\mathbf{2}}^n \Delta t s \label{eqn:ode_sol_crit}
\end{align}
where $\mathbf{g}_{\mathbf{o}}$ is the overdamped case, $\mathbf{g}_{\mathbf{u}}$ is the underdamped case, and $\mathbf{g}_{\mathbf{c}}$ is the critically damped case. As $\epsilon \to 0$, we obtain $\frac{e^{\epsilon} + e^{-\epsilon}}{2} \to 1$, $ \frac{e^{\epsilon} - e^{-\epsilon}}{2\epsilon} \to 1$, $\cos \epsilon \to 1$, $\frac{\sin\epsilon}{\epsilon} \to 1$; thus, $\mathbf{g}_{\mathbf{o}} \to \mathbf{g}_{\mathbf{c}}$ and $\mathbf{g}_{\mathbf{u}} \to \mathbf{g}_{\mathbf{c}}$. In all cases,
\begin{align}
    \boldsymbol{\gamma}_{\mathbf{1}}^n &= \mathbf{x}^{n} - \mathbf{p}(t^n)\label{eqn:ode_sol_gamma1}\\
    \boldsymbol{\gamma}_{\mathbf{2}}^n &= \dot{\mathbf{x}}^n +\frac{k_d}{2}\boldsymbol{\gamma}_{\mathbf{1}}^n - \dot{\mathbf{p}}(t^n). \label{eqn:ode_sol_gamma2}
\end{align}

\section{Learning the constitutive parameters}
\label{sec:optimization}

Given one or more temporal sequences $\left\{\boldsymbol\theta^1,\boldsymbol\theta^2,\hdots,\boldsymbol\theta^N\right\}$ and corresponding dynamic simulation or motion capture results $\big\{\mathbf{x}_D^1,\mathbf{x}_D^2,\hdots\\,\mathbf{x}_D^N \big\}$, we automatically learn constitutive parameters $k_s$ and $k_d$ for each spring. For each such temporal sequence, we create a loss function of the form
\begin{align}
   \mathcal{L} =  \sum_{n=1}^{N}\|\mathbf{x}(t^n) - \mathbf{x}_D^n\|_2^2 \label{eqn:loss}
\end{align}
where $\mathbf{x}(t^n)$ is determined as described in Section \ref{sec:dynamics}. When there is more than one temporal sequence, the loss function can simply be added together. Notably, the loss can be minimized separately for each particle in a highly parallel and efficient manner.
We use gradient descent, where initial guesses are obtained from a few iterations of a genetic algorithm \cite{holland1992adaptation}.

The gradient of $\mathcal{L}$ with respect to the parameters $k_d$ and $k_s$ requires the gradient of $\mathbf{x}(t^n)$ with respect to $k_d$ and $k_s$, i.e. $\frac{\partial \mathbf{x}}{\partial k_d}$ and $ \frac{\partial \mathbf{x}}{\partial k_s}$. From Equation \ref{eqn:ode_sol_general}, one can readily see that the chain rule takes the form
\begin{align}
    \frac{\partial \mathbf{x}}{\partial k_s} &= e^{-\frac{k_d}{2} \Delta t s } \frac{\partial \mathbf{g}}{\partial k_s} + \frac{\partial \mathbf{p}}{\partial k_s} \label{eqn:partial_x_ks}\\
    \frac{\partial \mathbf{x}}{\partial k_d} &= e^{-\frac{k_d}{2} \Delta t s } \frac{\partial \mathbf{g}}{\partial k_d} - \frac{\Delta t s}{2} e^{-\frac{k_d}{2} \Delta t s } \mathbf{g} +  \frac{\partial \mathbf{p}}{\partial k_d} \label{eqn:partial_x_kd}
\end{align}
where $\frac{\partial \mathbf{g}}{\partial k_s}$, $\frac{\partial \mathbf{g}}{\partial k_d}$, and $\mathbf{g}$ all vary based on $\epsilon$, i.e. based on whether $k_s$ and $k_d$ admit overdamping, underdamping, or critically damping. As we have seen (see Equation \ref{eqn:ode_sol_over}, \ref{eqn:ode_sol_under}, \ref{eqn:ode_sol_crit} and the discussion thereafter), $\mathbf{g}$ is continuous in the 2-dimensional $k_s$-$k_d$ phase space; however, one needs to carefully implement $\frac{\sin\epsilon}{\epsilon}$ and $ \frac{e^{\epsilon} - e^{-\epsilon}}{2\epsilon}$ to replace potentially spurious floating point divisions by the asymptotic result when $\epsilon$ is small. One can similarly show that $\frac{\partial \mathbf{g}}{\partial k_s}$ and $\frac{\partial \mathbf{g}}{\partial k_d}$ are continuous, and thus $\frac{\partial \mathbf{x}}{\partial k_s}$ and $\frac{\partial \mathbf{x}}{\partial k_d}$ are continuous.

To see that $\frac{\partial \mathbf{g}}{\partial k_s}$ and $\frac{\partial \mathbf{g}}{\partial k_d}$ are continuous, we expand them via the chain rule
\begin{align}
    \frac{\partial \mathbf{g}}{\partial k_s} &= \frac{\partial  \mathbf{g}}{\partial  \boldsymbol{\gamma}_{\mathbf{1}}^n}\frac{\partial \boldsymbol{\gamma}_{\mathbf{1}}^n}{\partial k_s} + \frac{\partial  \mathbf{g}}{\partial  \boldsymbol{\gamma}_{\mathbf{2}}^n}\frac{\partial \boldsymbol{\gamma}_{\mathbf{2}}^n}{\partial k_s} + \left(\frac{1}{\epsilon}\frac{\partial  \mathbf{g}}{\partial  \epsilon}\right)\left(\epsilon\frac{\partial \epsilon}{\partial k_s}\right) \label{eqn:partial_g_ks}\\
    \frac{\partial \mathbf{g}}{\partial k_d} &= \frac{\partial  \mathbf{g}}{\partial  \boldsymbol{\gamma}_{\mathbf{1}}^n}\frac{\partial \boldsymbol{\gamma}_{\mathbf{1}}^n}{\partial k_d} + \frac{\partial  \mathbf{g}}{\partial  \boldsymbol{\gamma}_{\mathbf{2}}^n}\frac{\partial \boldsymbol{\gamma}_{\mathbf{2}}^n}{\partial k_d} + \left(\frac{1}{\epsilon}\frac{\partial  \mathbf{g}}{\partial  \epsilon}\right)\left(\epsilon\frac{\partial \epsilon}{\partial k_d}\right) \label{eqn:partial_g_kd}
\end{align}
and note that $\frac{\partial  \mathbf{g}}{\partial  \boldsymbol{\gamma}_{\mathbf{1}}^n}$ and $\frac{\partial  \mathbf{g}}{\partial  \boldsymbol{\gamma}_{\mathbf{2}}^n}$ are continuous for the same reasons that $\mathbf{g}$ is. As can be seen in Equations \ref{eqn:ode_sol_gamma1} and \ref{eqn:ode_sol_gamma2}, $\frac{\partial \boldsymbol{\gamma}_{\mathbf{1}}^n}{\partial k_s}, \frac{\partial \boldsymbol{\gamma}_{\mathbf{1}}^n}{\partial k_d}, \frac{\partial \boldsymbol{\gamma}_{\mathbf{2}}^n}{\partial k_s}$, and $\frac{\partial \boldsymbol{\gamma}_{\mathbf{2}}^n}{\partial k_d}$ recursively depend on the prior interval via $\mathbf{x}^n$ and $\dot{\mathbf{x}}^n$ (and eventually the initial conditions) but add no new discontinuities of their own. We inserted $\frac{1}{\epsilon}$ and $\epsilon$ into the last term in both Equations \ref{eqn:partial_g_ks} and \ref{eqn:partial_g_kd} so that $\epsilon \frac{\partial \epsilon}{\partial k_s} = \mp \frac{1}{2} \Delta t^2 s^2$ and $\epsilon \frac{\partial \epsilon}{\partial k_d} = \pm \frac{1}{4} \Delta t^2 s^2 k_d$ are robust to compute (the $\mp$ and $\pm$ signs represent overdamping/underdamping respectively). Then, we write
\begin{align}
    \frac{1}{\epsilon}\frac{\partial \mathbf{g}_{\mathbf{o}}}{\partial \epsilon} &= \boldsymbol{\gamma}_{\mathbf{1}}^n \frac{e^{\epsilon} - e^{-\epsilon}}{2\epsilon}  +  \boldsymbol{\gamma}_{\mathbf{2}}^n \Delta t s \frac{(\epsilon-1)e^{\epsilon}+(\epsilon+1) e^{-\epsilon}}{2\epsilon^3} \label{eqn:partial_g_eps_over}\\
    \frac{1}{\epsilon}\frac{\partial \mathbf{g}_{\mathbf{u}}}{\partial \epsilon} &= -\left(\boldsymbol{\gamma}_{\mathbf{1}}^n \frac{\sin\epsilon}{\epsilon}  + \boldsymbol{\gamma}_{\mathbf{2}}^n \Delta t s \frac{\sin \epsilon - \epsilon \cos \epsilon}{\epsilon^3}\right) \label{eqn:partial_g_eps_under}
\end{align}
to identify two more functions that must be carefully implemented (as $\epsilon \to 0$, $\frac{(\epsilon-1)e^{\epsilon}+(\epsilon+1) e^{-\epsilon}}{2\epsilon^3}  \to \frac{1}{3}$ and $\frac{\sin\epsilon - \epsilon \cos \epsilon}{\epsilon^3} \to \frac{1}{3}$). The sign difference between Equation \ref{eqn:partial_g_eps_over} and \ref{eqn:partial_g_eps_under} matches that in $\epsilon\frac{\partial \epsilon}{\partial k_s}$ and $\epsilon\frac{\partial \epsilon}{\partial k_d}$ showing that both $\frac{\partial \mathbf{g}}{\partial \epsilon}\frac{\partial \epsilon}{\partial k_s}$ and $\frac{\partial \mathbf{g}}{\partial \epsilon}\frac{\partial \epsilon}{\partial k_d}$ are continuous.

Finally, it is worth noting that a 2-dimensional gradient cannot be computed on the codimension-1 curve associated with critically damping; however, taking the dot product of the continuous (between overdamping and underdamping) gradient with the tangent to the codimension-1 curve (and adjusting for either $k_s$ or $k_d$ parameterization) matches the derivative along the curve as expected.

\section{Results and Discussion}
\label{sec:result}

\begin{figure}
  \begin{subfigure}[b]{0.23\textwidth}
    \includegraphics[width=\textwidth]{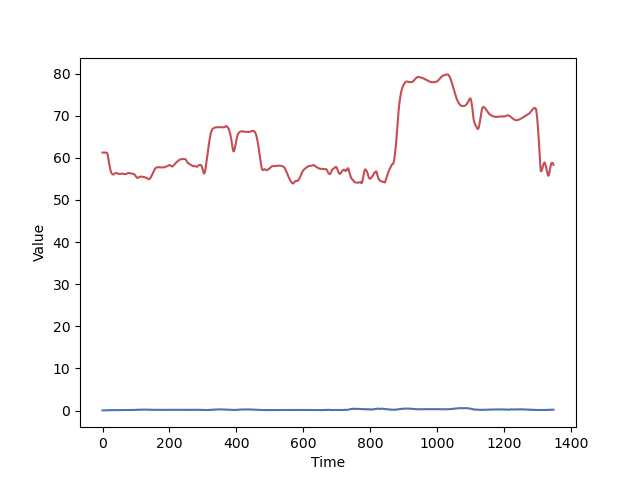}
    \caption{all vertices averaged}
    \label{fig:dyn_layer:all_1}
  \end{subfigure}
  \begin{subfigure}[b]{0.23\textwidth}
    \includegraphics[width=\textwidth]{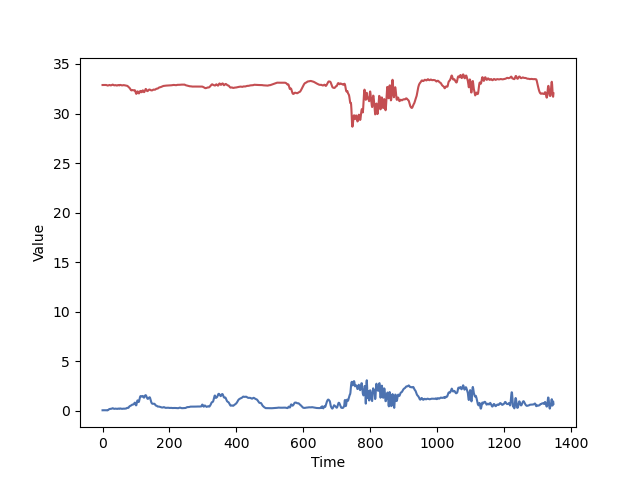}
    \caption{one vertex}
    \label{fig:dyn_layer:one_1}
  \end{subfigure}
  \begin{subfigure}[b]{0.23\textwidth}
    \includegraphics[width=\textwidth]{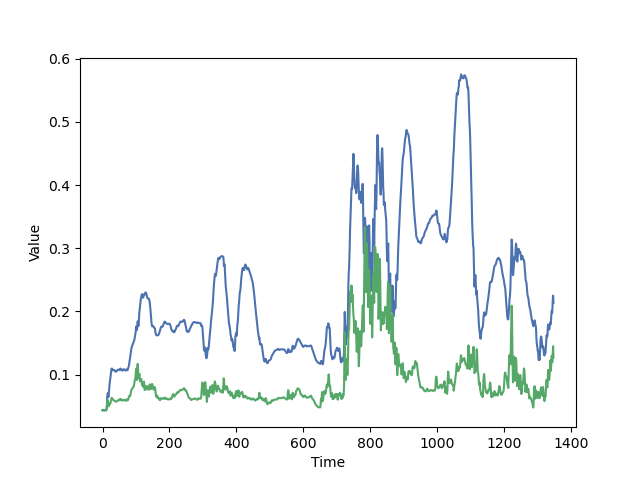}
    \caption{all vertices averged}
    \label{fig:dyn_layer:all_2}
  \end{subfigure}
  \begin{subfigure}[b]{0.23\textwidth}
    \includegraphics[width=\textwidth]{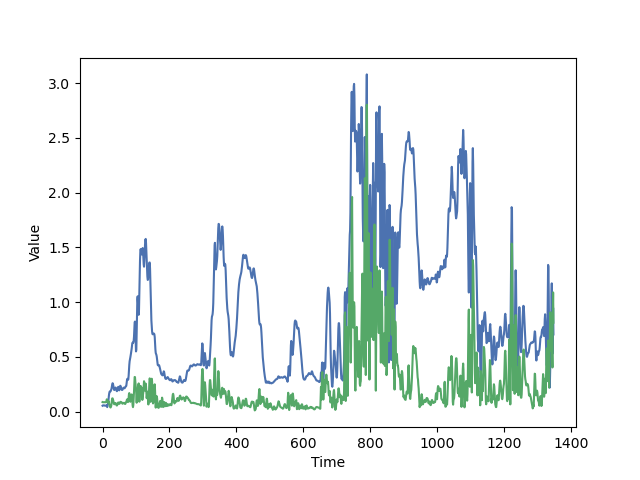}
    \caption{one vertex}
    \label{fig:dyn_layer:one_2}
  \end{subfigure}
  \begin{subfigure}[b]{0.23\textwidth}
    \includegraphics[width=\textwidth]{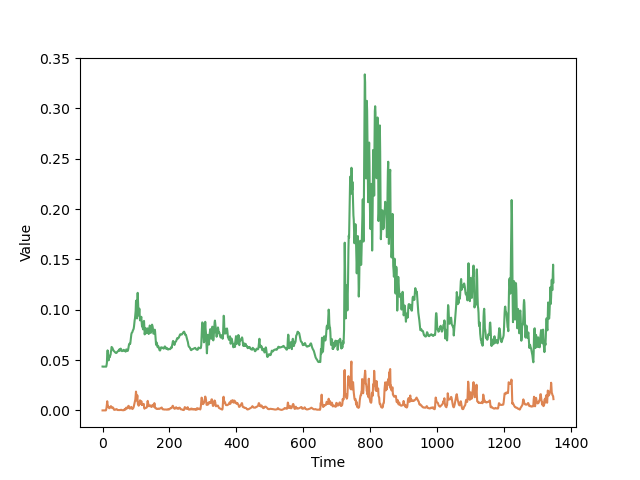}
    \caption{all vertices averged}
    \label{fig:dyn_layer:all_3}
  \end{subfigure}
  \begin{subfigure}[b]{0.23\textwidth}
    \includegraphics[width=\textwidth]{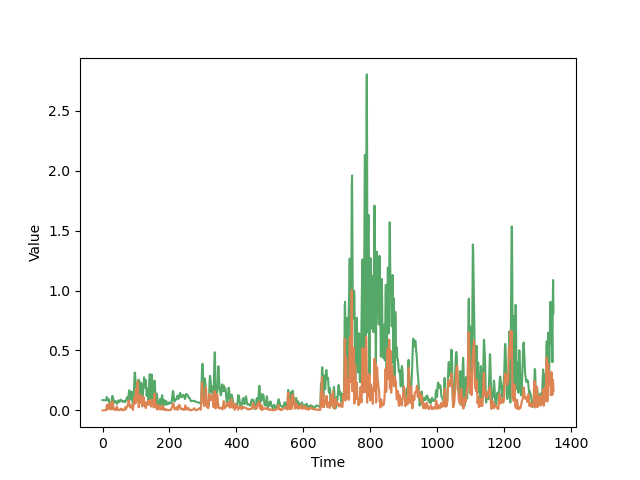}
    \caption{one vertex}
    \label{fig:dyn_layer:one_3}
  \end{subfigure}
  \caption{Red curve: $\ell_2$ norm of vertex positions in the pelvis coordinate system. Blue curve: $\ell_2$ norm of displacements from skinning to dynamics. Green curve: $\ell_2$ norm of displacements from QNN to dynamics. Orange curve: $\ell_2$ norm of displacements from QNN to zero-restlength springs.}
  \label{fig:dyn_layer}
\end{figure}

\begin{figure}
    \includegraphics[width=0.5\textwidth]{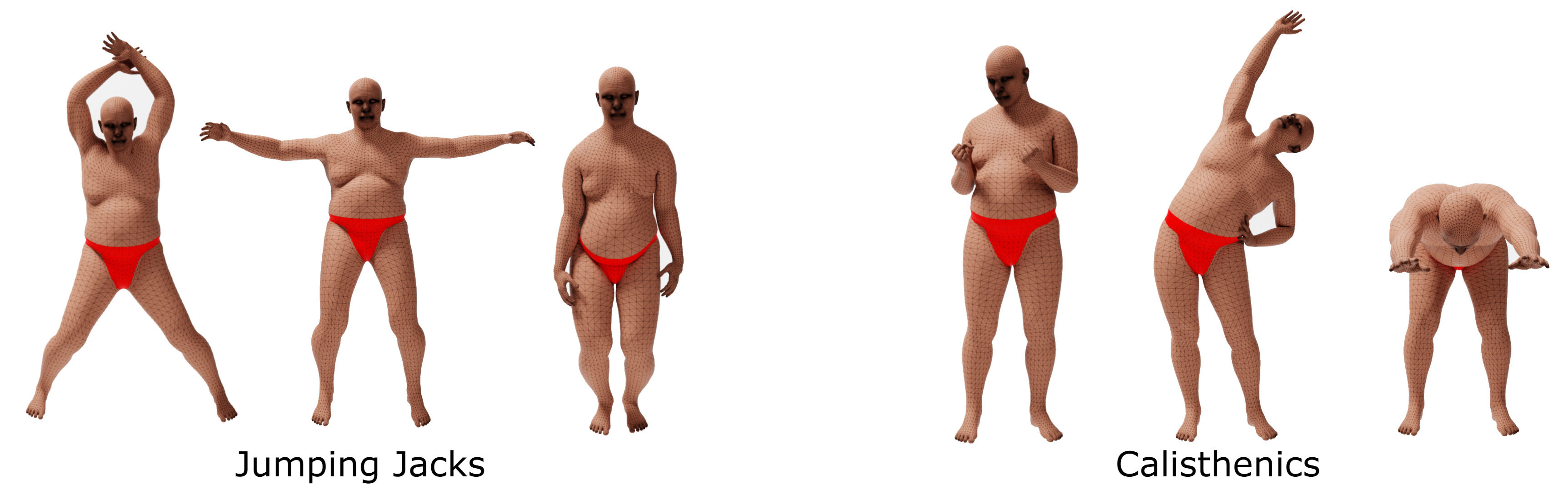}
    \caption{Dynamic simulation sequences used to learn zero-restlength spring constitutive parameters.}
    \label{fig:sim_data_example}
\end{figure}

Figure \ref{fig:dyn_layer} quantitatively illustrates how our approach alleviates the demand on the neural network for a particular dynamic simulation example (``calisthenics''). Figure \ref{fig:dyn_layer:all_1} shows the $\ell_2$ norm of the vertex positions (red curve) measured relative to a coordinate system whose origin is placed on the pelvis joint. Figure \ref{fig:dyn_layer:one_1} shows the same result for a single vertex on the belly. The $\ell_2$ norm of the displacements from the skinned result (blue curve) is vastly smaller (as shown in Figures \ref{fig:dyn_layer:all_1} and \ref{fig:dyn_layer:one_1}), indicating that most of this function is readily captured via skinning (a number of authors have utilized this approach \cite{pons2015dyna, santesteban2020softsmpl, seo2021dsnet, jin2020pixel, loper2015smpl}). In Figures \ref{fig:dyn_layer:all_2} and \ref{fig:dyn_layer:one_2}, we change the scale so that the blue curve can be more readily examined. In addition, we also plot the $\ell_2$ norm of the displacements from our QNN result (green curve) as the dynamics layer we want to approximate. This dynamics layer has a relatively small magnitude and low variance (comparably), which is readily approximated/learned based on a few dynamic simulations of training data. Figures \ref{fig:dyn_layer:all_3} and \ref{fig:dyn_layer:one_3} show the dynamics layer approximated by our zero-restlength springs (orange curve). Our approach captures the approximated shape, but with smaller magnitude, due to regularization. However, even with regularization, our method still outputs quite compelling dynamics (as can be seen in the supplementary video). 

As mentioned in Section \ref{sec:optimization}, we learn our spring constitutive parameters using a (surprisingly) small amount of ground truth simulation data. We obtain the dynamic simulation results $\left\{\mathbf{x}_D^1,\mathbf{x}_D^2, \hdots  ,   \mathbf{x}_D^N \right\}$ via backward Euler simulation. Figure \ref{fig:sim_data_example} shows examples of two dynamic simulation sequences (``jumping jacks'' and ``calisthenics'') we use to learn zero-restlength spring constitutive parameters. Note that any reasonable animation sequence with dynamics can be used, even motion capture data (see e.g. \cite{pons2015dyna}).
Although we use a dataset with 5000 data samples in order to train a robust QNN (see Section \ref{sec:qnn}), only a few dynamic simulation examples are required in order to learn zero-restlength spring constitutive parameters that generalize well to unseen animations. This also means that we only need to engineer the network architectures and hyperparameters for the configuration-only QNN, which is much easier than engineering a network that captures configuration transitions (see Section \ref{sec:intro} for the discussion about underfitting and overfitting of transition-based methods).

\subsection{Examples}

\begin{figure}
    \includegraphics[width=0.5\textwidth]{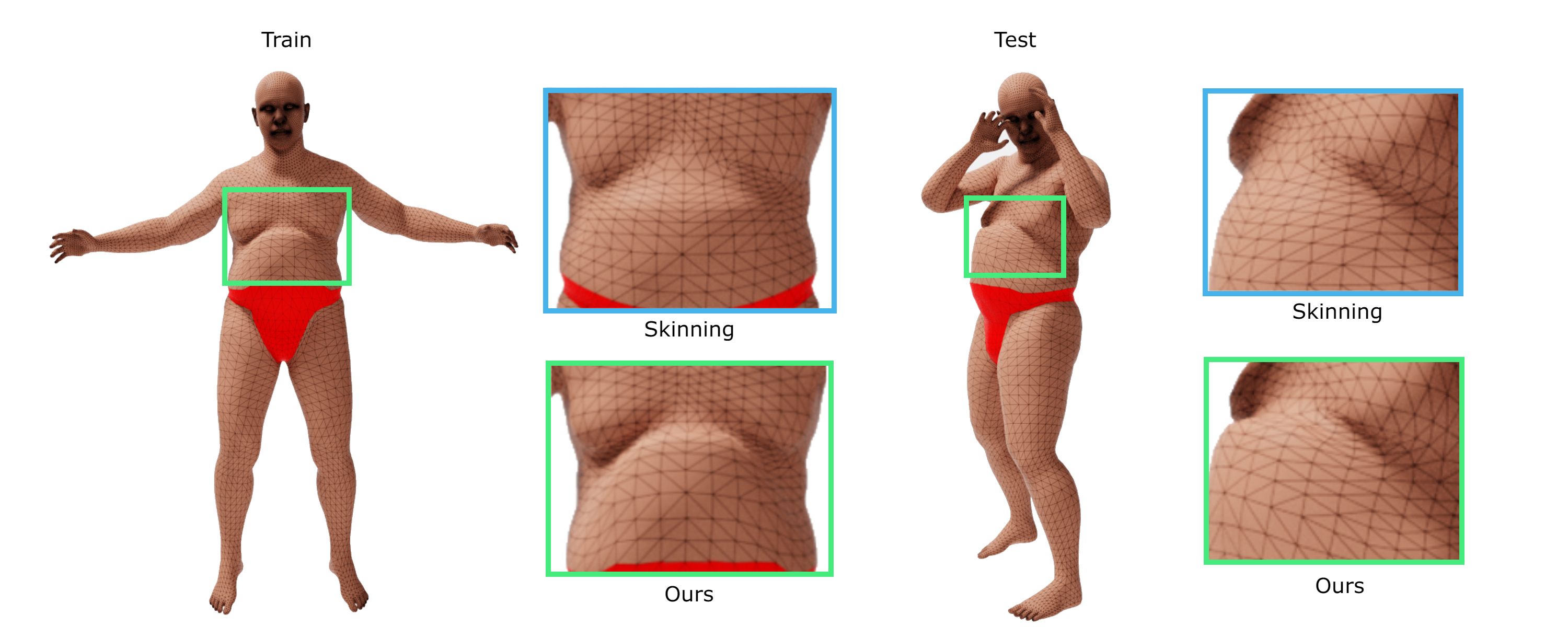}
    \caption{Comparison of our trained zero-restlength spring ballistic motion with the corresponding skinned result. Left: a motion sequence included in training. Right: a motion sequence not included in training. The ability to train on ``jumping jacks'' and generalize to ``shadow boxing'' would be impossible for a typical neural network approach.}
    \label{fig:dyn_example}
\end{figure}

Our analytic zero-restlength spring model generalizes very well to unseen animations and does not face severe underfitting or overfitting, which is common in machine learning methods if the network architecture is not carefully designed and trained on a plethora of data.
Figure \ref{fig:dyn_example} qualitatively shows two example frames comparing a skinning-only result with our analytic zero-restlength springs added on top of our QNN.
The frame on the left (``jumping jacks'') is taken from an animation sequence used in training while the frame on the right (``shadow boxing'') is taken from an animation sequence not used in training. In both examples, our method successfully recovers ballistic motion (e.g. in the belly).
Our method runs in real-time (30-90 fps, or even faster pending optimizations) and emulates the effects of accurate, but costly, dynamic backward Euler simulation remarkably well. We refer readers to our supplementary video for a compelling demonstration, particularly of the secondary inertial motion.

\begin{figure*}[t]
  \begin{subfigure}[b]{0.23\textwidth}
    \includegraphics[width=\textwidth]{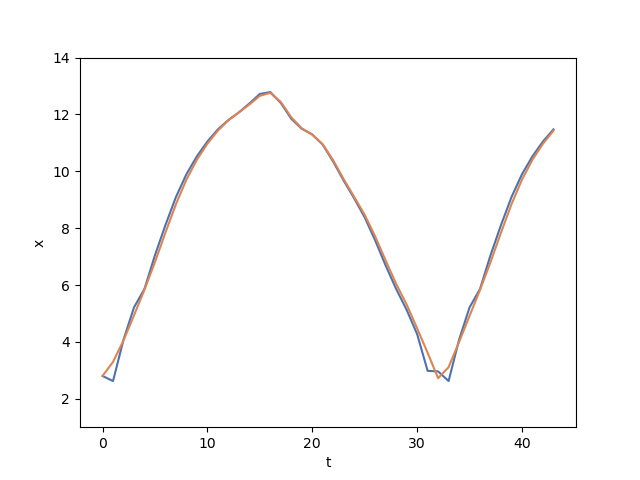}
    \label{fig:ransac:all_1}
  \end{subfigure}
  \begin{subfigure}[b]{0.23\textwidth}
    \includegraphics[width=\textwidth]{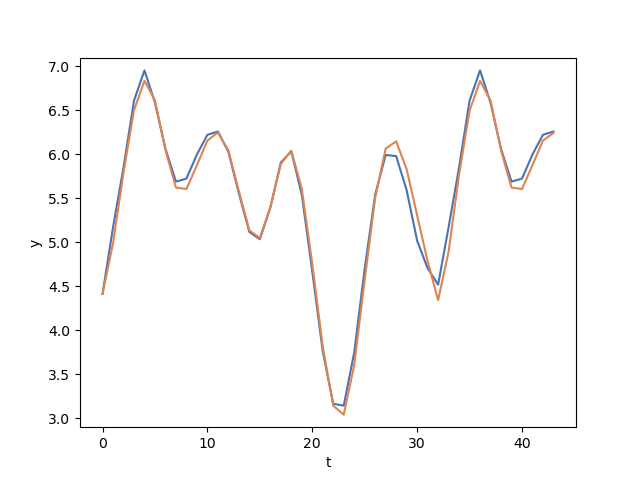}
    \label{fig:ransac:one_1}
  \end{subfigure}
  \begin{subfigure}[b]{0.23\textwidth}
    \includegraphics[width=\textwidth]{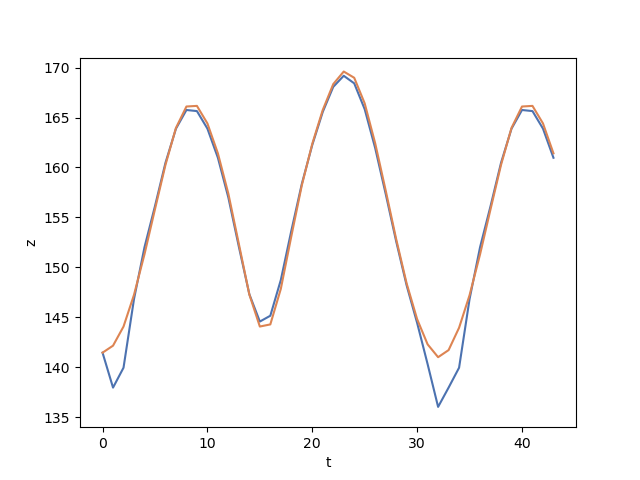}
    \label{fig:ransac:all_2}
  \end{subfigure}
  \begin{subfigure}[b]{0.23\textwidth}
    \includegraphics[width=\textwidth]{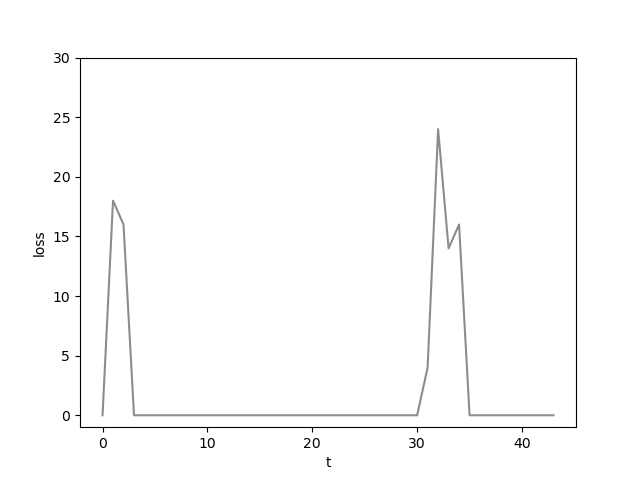}
    \label{fig:ransac:one_2}
  \end{subfigure}
  \begin{subfigure}[b]{0.23\textwidth}
    \includegraphics[width=\textwidth]{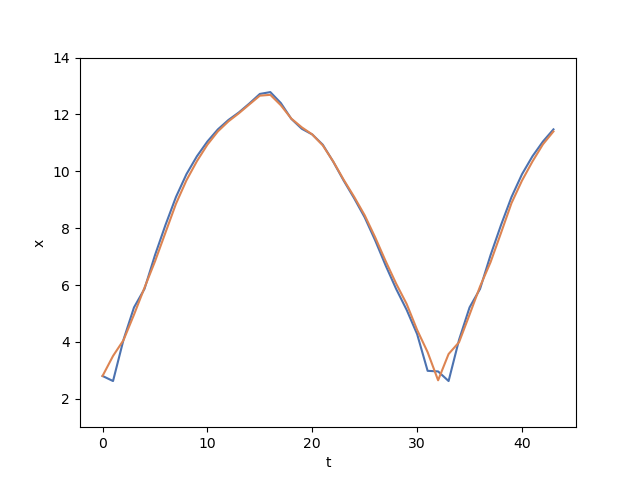}
    \caption{X axis}
    \label{fig:ransac:all_3}
  \end{subfigure}
  \begin{subfigure}[b]{0.23\textwidth}
    \includegraphics[width=\textwidth]{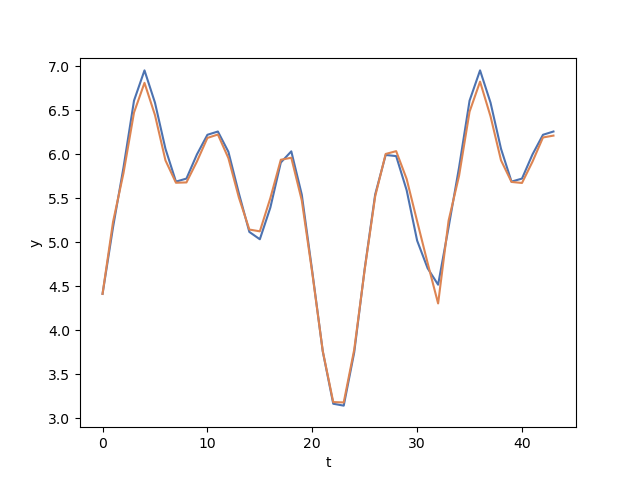}
    \caption{Y axis}
    \label{fig:ransac:one_3}
  \end{subfigure}
  \begin{subfigure}[b]{0.23\textwidth}
    \includegraphics[width=\textwidth]{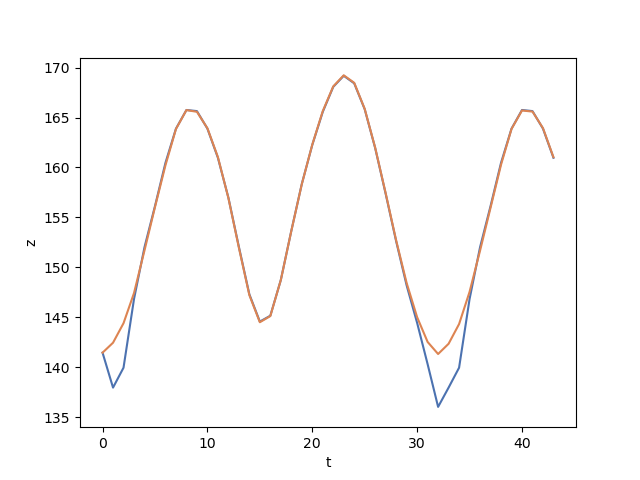}
    \caption{Z axis}
    \label{fig:ransac:all_3}
  \end{subfigure}
  \begin{subfigure}[b]{0.23\textwidth}
    \includegraphics[width=\textwidth]{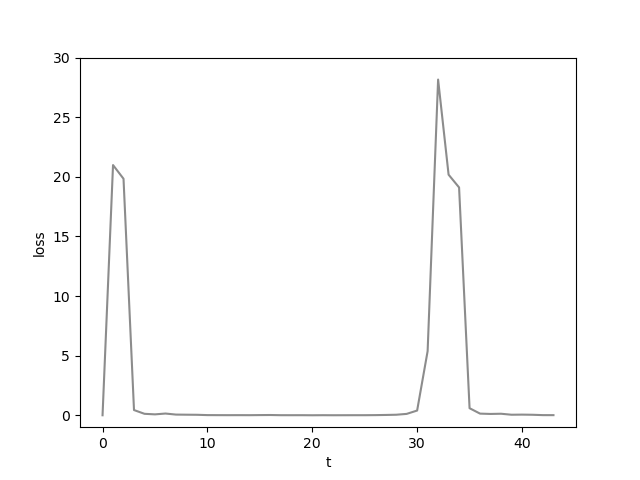}
    \caption{Loss}
    \label{fig:ransac:one_3}
  \end{subfigure}
  \caption{Robust training in the presence of simulation errors. Subfigures in columns (a)-(c) are per-axis trajectories of an example vertex in the jumping jack sequence. The backward Euler trajectory is shown in blue and our analytic zero-restlength spring trajectory is shown in orange. The high-frequencies in Frames 31-34 are caused by poorly converged dynamics in the presence of collisions. Subfigures in column (d) show the $\ell_2$ loss between the zero-restlength springs and backward Euler. The first row is the initial training result and the second row is the re-trained result with the 10\% highest-loss frames ignored. The second row more closely follows the backward Euler trajectory for the frames that don't have simulation errors.}
  \label{fig:ransac}
\end{figure*}

Full dynamic simulation is costly and prone to instabilities. Often this results in a few simulated frames with visible errors. To avoid such artifacts, we modify our training procedure to avoid overfitting to poorly converged frames (that would lead to poor generalization). See Figure \ref{fig:ransac}. We note that similar approaches are common in the computer vision community (see e.g. random sample consensus \cite{fischler1981random}).

\begin{figure}
    \includegraphics[width=0.5\textwidth]{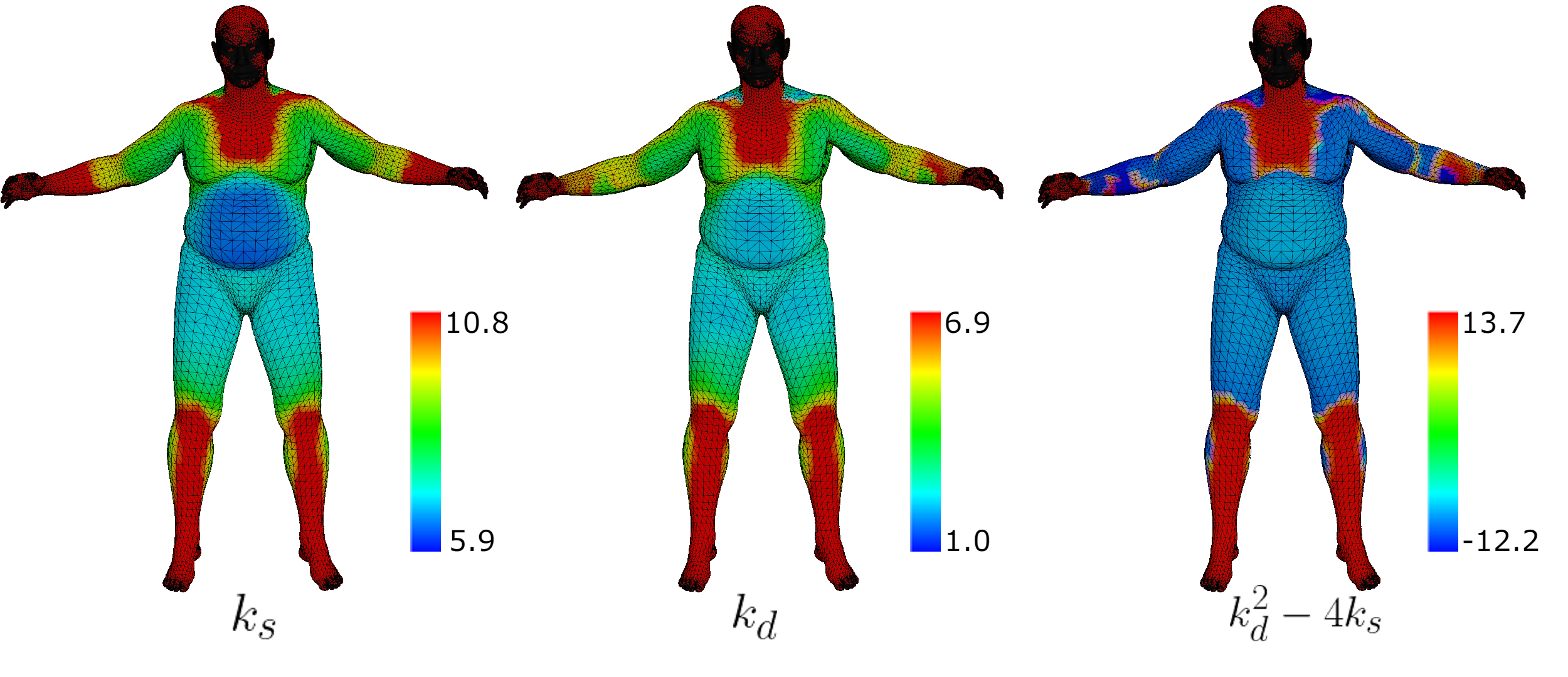}
    \caption{Heatmap visualization (logarithm scale) of stiffness $k_s$, damping $k_d$, and $k_d^2-4k_s$ which determines overdamping/underdamping, respectively. In heavily constrained regions the springs are stiffer and more overdamped, while in fleshy regions the springs are softer and more underdamped. Note that more constrained regions occur based on proximity to the bones used in the dynamic simulation training data (e.g. chest, forearms, shins, etc.). }
    \label{fig:interpretability}
\end{figure}

Figure \ref{fig:interpretability} shows a heatmap visualization of learned $k_s$, $k_d$ and the overdamping/ underdamping indicator $k_d^2 - 4k_s$, respectively. Note how symmetric our optimization result is, even if we optimize each particle. In regions where rigid motion dominates (e.g. hands, feet, head, etc.), the optimization results in overdamped springs with large stiffness. The code can be accelerated by replacing the constitutive parameters of all such springs with a single set of consitutive parameters. In regions where soft-tissue dynamics dominates (e.g. belly, thigh, etc.), the optimization results in underdamped springs with small stiffness. Since our optimization is per particle decoupled, it is easy to troubleshoot (if necessary).

As a final note, one could obviously add our zero-restlength springs on top of the skinned result directly; however, we obtained better results using our QNN to fix skinning artifacts due to volume loss and collision.

\section{Conclusion and Future Work}
We present an analytically integratable physics model that can recover dynamic modes in real-time. The main takeaway is that the problem can be separated into a configuration-only quasistatic layer and a transition-dependent dynamics layer, where the dynamics layer can be well approximated by a simple physics model. The constitutive parameters of the physics model can be robustly learned from only a few backward Euler simulation examples. In particular, determining $k_s$ and $k_d$ requires a gradient that can erroneously overflow/underflow near the critical damping manifold in $k_s$-$k_d$ phase space. We quite robustly addressed this by isolating non-dimensionalized functions that were trivially carefully implemented to obtain the correct asymptotic result in \textit{all} cases. For more discussions on both numerical and analytical issues with gradients, we refer the interested readers to \cite{johnson2022, metz2021gradients}.

\bibliographystyle{ACM-Reference-Format}
\bibliography{sample-base}
\end{document}